\documentclass[preprint,12pt]{elsarticle}

\usepackage{amssymb}
\usepackage{hyperref}
\usepackage{array, multirow}
\usepackage{graphicx}
\usepackage{caption}
\usepackage{algorithm}
\usepackage{algcompatible}
\usepackage{float}

\algnewcommand\algorithmicreturn{\textbf{return}}
\algnewcommand\RETURN{\State \algorithmicreturn}

\graphicspath{ {MIA_figures/} }

\journal{Medical Image Analysis}

\begin{document}

\begin{frontmatter}


\title{Quantification of Ultrasonic Texture heterogeneity via Volumetric Stochastic Modeling for Tissue Characterization}

\author[Affil1,Affil2]{Omar S ~Al-Kadi\corref{cor1}}
\author[Affil1]{Daniel YF Chung}
\author[Affil1]{Robert C Carlisle}
\author[Affil1]{Constantin C Coussios}
\author[Affil1]{J Alison Noble}
\address[Affil1]{Institute of Biomedical Engineering, Department of Engineering Science, University of Oxford, Oxford OX3 7DQ United Kingdom}
\address[Affil2]{King Abdullah II School for Information Technology, University of Jordan, Amman 11942, Jordan}
\cortext[cor1]{Corresponding Author: Omar S. Al-Kadi, E-mail: omar.al-kadi@eng.ox.ac.uk}

\begin{abstract}
Intensity variations in image texture can provide powerful quantitative information about physical properties of biological tissue. However, tissue patterns can vary according to the utilized imaging system and are intrinsically correlated to the scale of analysis. In the case of ultrasound, the Nakagami distribution is a general model of the ultrasonic backscattering envelope under various scattering conditions and densities where it can be employed for characterizing image texture, but the subtle intra-heterogeneities within a given mass are difficult to capture via this model as it works at a single spatial scale. This paper proposes a locally adaptive 3D multi-resolution Nakagami-based fractal feature descriptor that extends Nakagami-based texture analysis to accommodate subtle speckle spatial frequency tissue intensity variability in volumetric scans. Local textural fractal descriptors -- which are invariant to affine intensity changes -- are extracted from volumetric patches at different spatial resolutions from voxel lattice-based generated shape and scale Nakagami parameters. Using ultrasound radio-frequency datasets we found that after applying an adaptive fractal decomposition label transfer approach on top of the generated Nakagami voxels, tissue characterization results were superior to the state of art. Experimental results on real 3D ultrasonic pre-clinical and clinical datasets suggest that describing tumor intra-heterogeneity via this descriptor may facilitate improved prediction of therapy response and disease characterization.

\end{abstract}


\end{frontmatter}


\section{Introduction}

    Analysis of the local characteristic patterns of tissue texture can reveal subtle pathological features deemed important for clinical diagnosis. Spatial variation of textons quantified in terms of image ``surface roughness'' has been shown to reflect tumor functional heterogeneity, and to lead to a better understanding of disease state \cite{bae13, chl13, dvl12, alk08}.  However, sub-voxel resolution complex and higher order textural features can be difficult to discern by simple observation. These texture signatures may convey significant information about disease progression or regression. However, quantifying these subtle signatures in ultrasound images is challenging.

    Our motivation stems from a clinical need to improve the diagnosis and therapy of liver cancer. Approximately 100,000 patients are diagnosed each year with primary liver cancers in the United States and Europe \cite{cuk14, acs14}. When this is compared against worldwide statistics, liver cancer is even more common in developing countries \cite{who12}. Although it the sixth most common cancer in the world \cite{fry13}, incidence varies across the world, and it is the most cancer type in some developing countries \cite{prk14}. Surgery is considered the only curative treatment; however, this is not suitable in the majority of cases due to co-morbidity, extent or location of the cancer, with chemotherapy forming the mainstay of treatment in these patients. Chemotherapy can have significant side effects, and may not be effective in all cases. Development of monitoring techniques during the course of chemotherapy may permit dose adjustment in responders to minimize side effects, while alternative treatments could be offered to non-responders. Current monitoring techniques rely on computed tomography and magnetic resonance imaging, with frequency limited by the potential damage from ionizing radiation and cost consideration respectively. Despite the difficulties of using ultrasound for monitoring disease (e.g. operator dependent, poorly reproducible and non-standardize), it is a technique that is known to be rapid, relatively inexpensive, readily available, with no exposure to ionizing radiation, making it ideal for frequent monitoring of liver tumors during a course of treatment.
    
    Given the advantages of ultrasound, analyzing tissue speckle from a single resolution perspective is limiting, as substantial information that could assist tumor tissue characterization can be hidden at sub-voxel resolution. This is true for the smaller necrotic or functionally low-activity regions that exhibit a hyperechogenic appearance compared to healthy tissue \cite{uta11}. We hypothesize that the difference in echogenicity of the tumor speckle texture can be exploited as an indicator of disease responsiveness to treatment \cite{czr13}. Nevertheless, the functionally low-activity regions within the tumor texture are relatively small, especially in the early sessions of tumor chemotherapy treatment. They also tend to have low intensity contrast compared to the aggressive or functionally active background of the remaining tumor. Identification of subtle changes in these regions based on visual assessment of the intensity alone can be challenging.
    
\indent Our approach is motivated by four observations:
\begin{itemize}
\item {\em Tumors are heterogeneous}: most previous work has accounted for functionally active malignant regions rather than the peripheral low activity necrotic regions which may additionally provide key information on disease progression or regression. These subtle variations and deviations within the speckle tissue texture were deemed too chaotic to be characterized in \cite{lru14}, but are important for understanding disease state; 
\item {\em Heterogeneity suggests using a multi-resolution texture analysis}: a carefully designed multi-resolution approach which is visually discriminative and geometrically informative could reveal small speckle changes and is better suited to describe the mixture distribution complexity that underpins a heterogeneity model; 
\item {\em Fractal analysis is well-suited to this problem}: conventional energy-based wavelet decompositions are susceptible to local intensity distribution variations; the fractal signatures used in our approach, derived from the wavelet representation sub-bands related to physiological properties of texture surface roughness are not; finally, 
\item {\em Analysis should be three-dimensional}: performing a 3D texture analysis based on a volumetric Nakagami modeling could facilitate a more reliable estimate of the Nakagami parameters, where the 3D location of each voxel provides a better localization of speckle distribution mixtures. 
\end{itemize}

\begin{figure}
\includegraphics [width=\columnwidth]{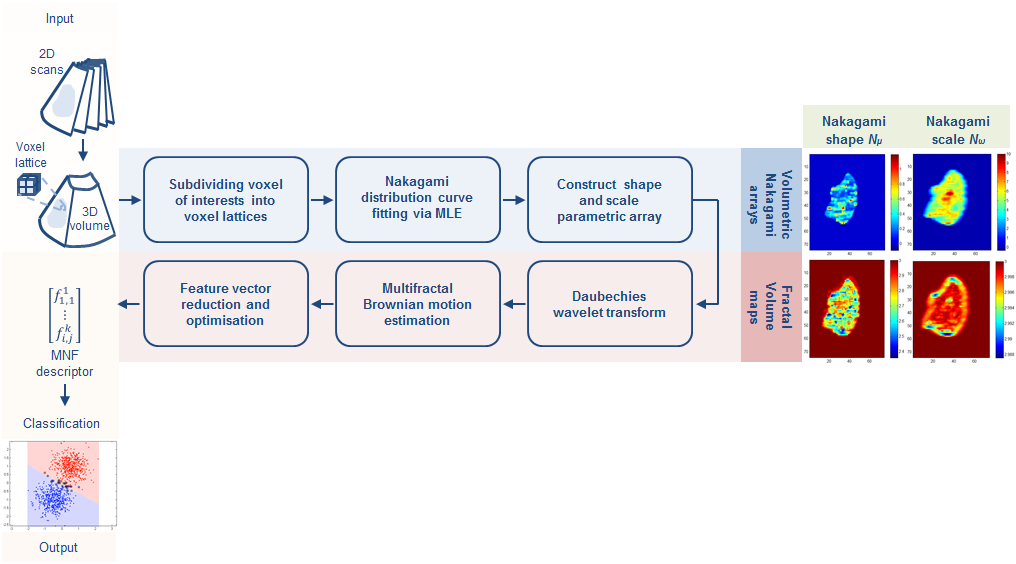}
\caption{3D multifractal Nakagami feature descriptor algorithm design for ultrasonic tissue characterization introduced in this paper.}
\label{fig:MNF-pipeline}
\end{figure}

In this work, a novel multifractal Nakagami-based volumetric feature descriptor that is invariant to local speckle attenuation changes is proposed. A pipeline summarizing the stages of our approach is illustrated in Fig.~\ref{fig:MNF-pipeline}. It is postulated that fractal tissue characteristics locally derived from 3D textural tumor patterns at several scales and from the RF envelope of the ultrasound backscattered volumes can assist in attaining descriptive features that relate to underlying biological structure. These tissue textural fractal characteristics tend to change in cases of therapeutic response, providing an attractive indicator for disease response to treatment during chemotherapy.   \\
     \indent This paper is organized as follows. State of the art and challenges associated with characterizing speckle tissue texture heterogeneity are summarized in Sections 2 and 3, followed by a detailed explanation of the proposed 3D multifractal Nakagami-based feature descriptor in Section 4. Sections 5 and 6 present the experimental results and discuss the potential significance of the work. The paper concludes in Section 7. 

\section{Related work}    
    One of the effective statistical techniques used for modeling various backsc\-attering conditions in medical ultrasound is the Nakagami distribution. This probabilistic distribution is known for its analytical simplicity and effective modeling of dense scatterers, accounting for amplitude and spacing, and can be reduced to a Rayleigh distribution under certain assumptions of scatterer density and number \cite{skr00}. Shankar \textit{et al.} first proposed the Nakagami distribution for characterizing conditions ranging from pre- to post-Rayleigh existing in ultrasound images, and later for modeling the radio-frequency (RF) envelope of the ultrasound backscattered signal in characterizing B-mode breast masses \cite{skr01}. Others have attempted to tackle the issue of accurate estimation of the Nakagami distribution. For instance \cite{lru11} employed Gamma kernel density estimation to compute a smooth estimate of a distribution within small windows of B-mode ultrasound images, but the mixture of distributions occurring at the boundaries between structures was not accounted for. The impact of morphological parameters and tumor structures on the Nakagami parameters statistics were analyzed in \cite{lru14}. A limitation observed was that there was a need for a robust algorithm to compute the Nakagami parametric images that better delineate the structures and the context in and around the tumor. Characterizing homogeneous tissues via improving the smoothness of the Nakagami parametric images was shown in \cite{tsi14}. The technique relies on summing and averaging the Nakagami images formed using sliding windows with varying window sizes related to the transducer pulse length. However, a relatively large window size (required for stabilization) may affect the reliability of the estimated Nakagami parameters, and hence degrade the spatial resolution of the resulting Nakagami image. 
     
    The Nakagami distribution has been further employed as an image feature in various image analysis contexts. For instance, five contour features and the two Nakagami parameters were used for classification of benign and malignant breast tumors in \cite{tsi10_tmi}. In a subsequent work malignant tumors were shown to be more pre-Rayleigh distributed than those from benign counterparts \cite{tsi10_umb}; however, the calculation of the average intensity value in the Nakagami image makes it susceptible to spatial frequency intensity variability. Further, that particular technique was optimized for 2D ultrasound images which may not reliably represent heterogeneous distributions of scatterers (or speckle) encountered within a tumor volume. A Random Forest based solution to learn tissue-specific ultrasonic backscattering and a signal confidence for predicting heterogeneous composition in atherosclerotic plaques was proposed in \cite{sht14}. That technique was developed for intravascular ultrasound and risk assessment of plaque rupture \cite{zhu02}. Necrotic core was not considered in that method. Finally, \cite{bhl09, kln11} describe a Markov random field model combined with Nakagami distribution estimation to differentiate malignant melanoma from normal tissue. However it was found that the estimated scale model parameter was highly sensitive to image quality, and hence subtle variations could go unnoticed. For an overview of ultrasound tissue characterization we refer the reader to \cite{nbl10}. Previous work on ultrasound texture analysis of tumors has considered both global and local non-uniformity quantification of the tumor texture at only a single analysis scale. Herein we are primarily concerned with tumor intra-heterogeneity (i.e. micro-structures within the tumor speckle texture) which is more challenging.
    
\section{Challenges in ultrasonic speckle texture characterization}

\begin{figure}
		\includegraphics[width=\columnwidth]{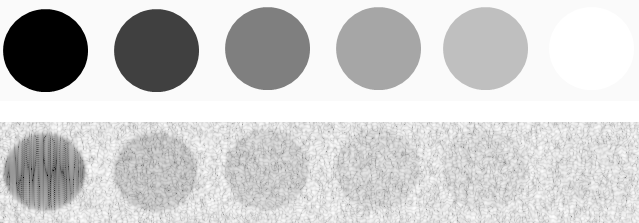}
		\caption{Six ultrasound hypoechoic to hyperechoic gray scale target phantoms having 8 mm diameter and 4 cm depth and corresponding simulated B-mode image representing a varying intensity from hypoechoic, -6, -3, +3, +6 dB, and hyperechoic, respectively.}
		\label{fig:ultrasound-simulation}
		\end{figure}

Speckle is a granular-shape stochastic pattern which appears in an image resulting from the scattering of an RF incident signal on an object~\cite{sch11}. The spacing and localization of the scatterers in the scanned object structure contribute to the local variation and distribution of the recorded texture pattern. However the characteristic interference patterns, known as speckle, produce an overall reduction in global image contrast~\cite{nbl10}. As a consequence, the boundaries separating different structures are less well defined, increasing difficulty in delineating regions of interest with a resultant increase in inter-and intra-observer variability for tumor detection. \\
    \indent A means to mitigate against effects such as the beam-tissue physical interaction and other acquisition factors is to characterize the objects via their speckle textural properties \cite{dgr03, mdb03, sdg13}. Textons or texels (texture elements) which are the fundamental components of texture that collectively form the observed speckle pattern texture do not directly correspond to the underlying structure; however, the local intensity textural pattern can reflect the local echogenicity of the underlying scatterers \cite{ads06}, see Fig.~\ref{fig:ultrasound-simulation}. This is due to the stochastic nature of the speckle pattern. Viewing the structure locally as a collective texton structure can give information about the underlying scatterer behavior. We hypothesize here this may lead to an improvement in internal structure delineation, and hence tumor characterization.
		
		\begin{figure} [t]
		\includegraphics[width=\columnwidth]{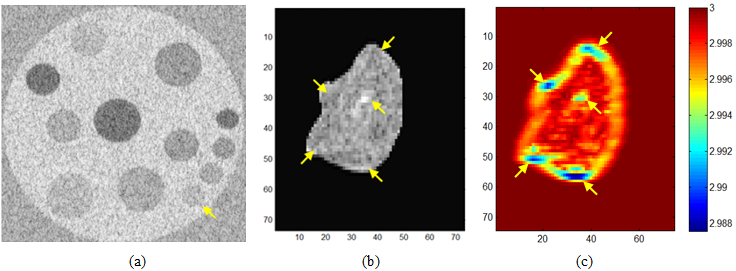}
		\caption{(a) Simulated ultrasound B-mode image following the method in \cite{bam80} showing different 4 cm depth of 4, 6 and 8 mm diameter gray scale target phantoms ranging from -6, -3, +3 and +6 dB varying intensity, (b) a real ultrasound B-mode volume of interest of a liver tumor with corresponding \textit{fractal slice map} in (c) -- estimated from the RF envelope of the ultrasound backscattered signal -- indicating the subtle low-activity regions.}
		\label{fig:ultrasound-simulation2}
		\end{figure}
		
\indent Fig.~\ref{fig:ultrasound-simulation2} (a) shows a simulated lesion phantom having three different sizes at a fixed depth and with four levels of intensity contrast variability. The small round simulated hyperechoic region (marked with an arrow) resembles the functionally low-activity regions on real B-mode ultrasound images; appearing subtle and challenging to identify. Note that the situation would be even more complicated in real B-mode ultrasound tissue where the functionally active background of the aggressive tumor tissue would not be as uniform as in this example, and the non-aggressive regions do not usually have a constant intensity distribution, see Fig.~\ref{fig:ultrasound-simulation2} (b). 

The tissue characterization solution, as discussed in the paper, is to use a multi-resolution approach that highlights higher order statistical features of the RF envelope. Such features could go unnoticed in B-mode ultrasound and an experienced observer could struggle to identify subtle interval changes in these important texture features. \\
    \indent Fractals and wavelet packet analysis provide effective ways to break down statistical complexity to distinguish between different texture regions, where the invariance to affine speckle intensity changes for the former and the high sensitivity to local features for the latter facilitates effective texture discrimination \cite{alk09}. Furthermore, according to the uncertainty principle, the wavelet packets can achieve an optimal joint spatial-frequency localization, i.e. simultaneously maintain a good boundary accuracy and frequency response~\cite{rgn05}, and the estimated fractal dimension can give a quantitative assessment of the surface roughness \cite{alk08, lop09}. Finally, simultaneous macro and micro scale tumor texture analysis provides a more complete characterization of dense and sparse textural regions within a tumor volume of interest. As demonstrated later, this progressive refinement process optimizes characterization by giving a better fit to the underlying tumor speckle texture.

\section{Methodology}
Our goal is to derive a locally-based feature signature based on volumetric generated Nakagami shape and scale parametric voxel lattices, and subsequently to perform an intensity-invariant texture analysis at various spatial resolutions for tissue characterization. This allows us to perform a more complete characterization of tumor texture at the optimal resolution scales compared to single or mono-resolution approaches \cite{alk09}. The proposed volumetric dense-to-sparse approach can break-down the speckle complexity and provide a robust estimation of model parameters, while having the advantage of simultaneously localizing both large high-contrast and small low-contrast structures at low and high spatial resolution levels. 

\subsection{Nakagami probabilistic distribution}

The Nakagami distribution \textit{N}(\textit{x}) is an analytically simple distribution that has been proposed as a general model for the ultrasonic backscattered envelope under all scattering conditions and scatterer densities \cite{skr00}. This distribution has the density function:

\begin{equation}
N(x|\mu,\omega)=2\left(\frac{\mu}{\omega}\right)^{\mu}\frac{1}{\Gamma\left(\mu\right)} x^{\left(2\mu-1\right)} e^{-\frac{\mu}{\omega}x^{2}} ,  \quad \forall x \in \Re \geq 0
\end{equation}

\noindent where $x$ is the envelope of the RF signal, with the shape of the distribution defined by the $\mu$ parameter corresponding to the local concentration of scatterers, and the local backscattered energy represented by the scale parameter $\omega$ $>$ 0, for $x$ $>$ 0, and $\Gamma\left(\cdot\right)$ is the Gamma function. If $x$ has a Nakagami distribution $N$ with parameters $\mu$ and $\omega$, then $x^2$ has a Gamma distribution $\Gamma$ with shape $\mu$ and scale (energy) parameter $\omega$/$\mu$. \\
    \indent The Nakagami distribution can model various backscattering conditions in medical ultrasound. By varying $\mu$, the envelope statistics range from pre-Rayleigh $\left(\mu < 1\right)$, Rayleigh $\left(0 < \mu < 0.5\right)$, and to post-Rayleigh $\left(\mu > 1 \right)$.  The Nakagami parameters are generally estimated by the $2^{nd}$ and $4^{th}$ order moments, where given $x$ is the ultrasonic backscattered envelope and $E\left(\cdot\right)$ denotes the statistical mean, the two Nakagami parameters can be calculated as:
\begin{equation}
	\omega = E\left(x^{2}\right) , and \quad \mu = \frac{E\left(x^{2}\right)^{2}}{Var\left(x^{2}\right)} = \frac{E\left(x^{2}\right)^{2}}{E\left(x^{4}\right) - E\left(x^{2}\right)^{2} .}
\end{equation}

		
\subsection{Volumetric multi-scale Nakagami modeling}

\begin{figure}
		\begin{center}
		\includegraphics[scale = 0.85]{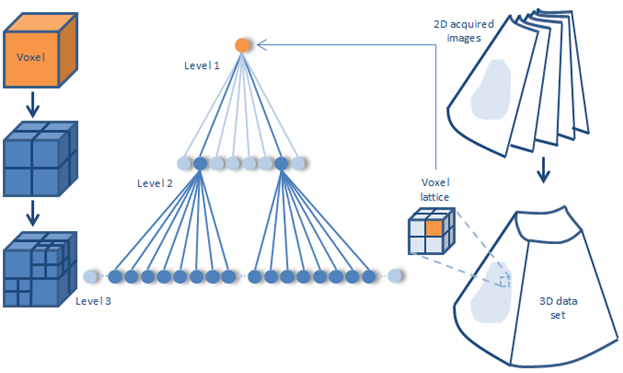}
		\end{center}
		\caption{Multiresolution volumetric modeling showing the decomposition up to 3 hierarchical levels by recursive subdivision of volume into octants voxels (left) and corresponding decomposition tree (middle).}
		\label{fig:voxel-decomposition}
		\end{figure}

A 3D feature signature that operates locally is defined by having each volume $V$ consisting of $z$ acquired slices $\left\{I_i: i = 1,\ldots, z\right\}$ subdivided into voxel lattices $v^{i}$, each having a defined size of $m$ and $n$, where $v = \left\{v_{kl}|k,l \in V\right\}$ for $k = 1,\ldots,m$, $l = 1,\ldots,n$, such that $\bigcup_{kl}v_{kl}=V$. For each $v^{i}$ we assume that for a scaling factor $r$ at a specific spatial scale $s$, the scaled voxel intensity lattice values $v^i_{klr}$ of the RF envelope amplitude $A_{klr}$ such that $A_{klr} = \left(v_{klr}\right)_{r\in R_s}$, where the different possible resolution levels $Rs: r = 1,\ldots,s,\ldots,j$ reaching to the maximum level $j$ represent a stochastic pattern, and the envelope amplitude of the scales $r$ of $v^i_{klr}$ follows a Nakagami distribution. Given the large number of voxel samples to analyze and the known family of probability distributions, the maximum likelihood estimators would tend to have a higher probability of being close to the quantities to be estimated and more often unbiased as compared to moments-based estimation \cite{chg01}, therefore the associated shape and scale parameters were estimated via maximum likelihood estimation (MLE) by operating on each voxel lattice region and at different scales. The maximum likelihood estimate $\hat{\theta} \left(v\right)$ for a density function $f\left(v^1_{111},\ldots,v^z_{mnj} | \theta\right)$ when $\theta$ is a vector of parameters for the Nakagami distribution family $\Theta$, estimates the most probable parameters $\hat{\theta}\left(v\right) = arg max_\theta \: D\left(\theta|v^1_{111},\ldots,v^z_{mnj}\right)$, where $D\left(\theta |v\right) = f\left(v|\theta\right), \! \theta \in \Theta$ is the score function. Having generated voxel-based Nakagami parameters, 3D wavelet packet Daubechies analysis \cite{mlt99} can be applied at multiple scales. Namely:

\begin{equation}
W_\varphi \left(t_0, x, y, z\right) = \frac{1}{\sqrt{mnj}} \sum^{j-1}_{r=0} \sum^{n-1}_{l=0} \sum^{m-1}_{k=0} v_{klr}\varphi_{klr}\left(t_0, x, y, z\right) ,
\end{equation}

\begin{equation}
W^i_\psi \left(t, x, y, z\right) = \frac{1}{\sqrt{mnj}} \sum^{j-1}_{r=0} \sum^{n-1}_{l=0} \sum^{m-1}_{k=0} v_{klr}\psi^i_{klr}\left(t, x, y, z\right) ,
\end{equation}
\\
\noindent where $v_{klr} \in L^2\left(\Re\right)$ is relative to scaling $\varphi_{klr}$ and wavelet function $\psi_{klr}$ and $W_\varphi\left(t_0, m, n\right)$ defines an approximation of $v_{klr}$ at scale $t_0$, and $W^i_\psi\left(t, m, n\right)$ coefficients add horizontal, vertical and diagonal details for scales $t \geq t_0$. The Daubechies wavelet family can account for self-similarity and signal discontinuities, making it one of the most useful wavelets for characterizing signals exhibiting fractal patterns \cite{dub90}. In our case an orthogonal 8-tap Daubechies filter was used to obtain the wavelet packets by expanding the basis having the most significant fractal signature rather than energy.\\
    \indent An octant wavelet transform depends mainly on the scaling $h_0\left(k\right)$ and wavelet $h_1\left(k\right)$ filters for image decomposition, and one does not need to express the $\varphi_{klr}$ and $\psi_{klr}$ in their explicit form. The decomposition process can be viewed as passing the signal through a pair of lowpass $\left(L\right)$ and highpass $\left(H\right)$ filters, also known as \textit{quadrature mirror} filters, having impulse responses $\tilde{h}_0\left(k\right)$ and $\tilde{h}_1\left(k\right)$, while holding the size of the transformed image the same as the original image as we are applying an overcomplete wavelet representation; hence giving a better representation of the texture characteristics at each decomposition. The impulse responses of $L$ and $H$ are defined as $\tilde{h}_0\left(a\right) = h_0\left(-a\right)$ and $\tilde{h}_1\left(a\right) = h_1\left(-a\right)$ for scaling parameter $a$, and $\tilde{h}_0\left(b\right) = h_0\left(-b\right)$ and $\tilde{h}_1\left(b\right) = h_1\left(-b\right)$ for translation parameter $b$, where $a, b \in Z$. The decomposition is performed recursively on the output of $\tilde{h}_0\left(a\right)$, $\tilde{h}_1\left(a\right)$ and $\tilde{h}_0\left(b\right)$, $\tilde{h}_1\left(b\right)$. Hence, the 3-D wavelet (or octant wavelet packet) can be expressed by the tensor product of the wavelet basis functions along the horizontal, vertical and depth directions. The corresponding filter coefficients can be recursively decomposed by a factor of eight as illustrated in Fig.~\ref{fig:voxel-decomposition} and expressed in (5), with subscripts indicating the low and high pass filtering characteristics in the $m$, $n$ and $j$ directions:
		\begin{eqnarray} 
		h_{LLL}\left(a,b\right) = h_0\left(a\right) h_0\left(a\right) h_0\left(b\right) , & h_{LHL}\left(a,b\right) = h_0\left(b\right)h_1\left(a\right)h_0\left(b\right) , \nonumber \\
		h_{LLH}\left(a,b\right) = h_0\left(a\right)h_0\left(a\right)h_1\left(b\right) , & h_{LHH}\left(a,b\right) = h_0\left(b\right)h_1\left(a\right)h_1\left(b\right) , \nonumber \\
		h_{HLL}\left(a,b\right) = h_1\left(b\right)h_0\left(a\right)h_0\left(b\right) , & h_{HHL}\left(a,b\right) = h_1\left(a\right)h_1\left(a\right)h_0\left(b\right)  ,\nonumber \\
		h_{HLH}\left(a,b\right) = h_1\left(b\right)h_0\left(a\right)h_1\left(b\right) , & h_{HHH}\left(a,b\right) = h_1\left(a\right)h_1\left(a\right)h_1\left(b\right)  .
		\end{eqnarray}
		
		\indent By decomposing the approximation coefficients of the signal as well, the wavelet transform can be extended in the middle and high frequency channels, providing a more complete partitioning of the spatial-frequency domain, which is known as the octant wavelet packet transform \cite{coi92}. As the textural information about the structural arrangement of surfaces and their relationship to the surrounding neighborhood is spread across the frequency sub-bands, most of the important discriminant features related to the structure terminations and endpoints of surface edges will have a stronger response in higher frequencies \cite{alk09}. Thereby, this gives an equal opportunity for investigating descriptors of textural features prevailing in the middle and high frequency bands. 
		
\begin{figure}
		\begin{center}
		\includegraphics[scale = 0.85]{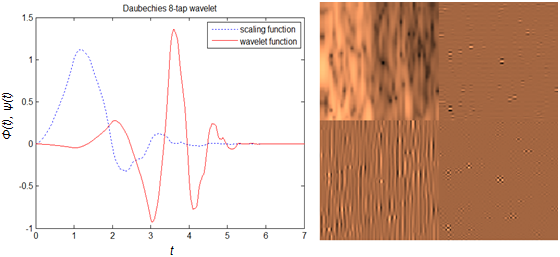}
		\end{center}
		\caption{Normalized Daubechies' orthogonal wavelet showing scaling (father) $\phi(t)$ and wavelet (mother) $\psi(t)$ functions with 4 vanishing moments, and corresponding first level Daubechies wavelet decomposition for a liver tumor volume of interest showing from left to right the approximation, horizontal, vertical, and diagonal coefficients, respectively.}
		\label{fig:wavelets}
		\end{figure}

    \indent From a pattern recognition perspective, the selection of the most suitable wavelet is associated with the understanding of the tissue textural properties and synthesis wavelet. Wavelet analysis using Daubechies wavelet basis functions can achieve a good spatial-frequency localization by having narrow high and wide low frequencies simultaneously. With the increasing number of zero or vanishing moments -- which are half the number of filter taps $N$ -- this can give a sparse representation for a large class of signal types. Also the Daubechies orthogonal wavelet family consists of purposefully designed filters which account for self-similarity and signal discontinuities, making them one of the most useful wavelets for characterizing signals exhibiting fractal patterns. Besides, they are also considered to be sensitive in recognizing fine characteristic structures, and its application of overlapping windows, unlike other wavelets such as the Haar wavelet, facilitates the capture of all high frequency changes easily \cite{mlt99}. As our work is concerned with the estimation of texture surface roughness from a fractal dimension perspective, the choice of this wavelet is more suitable than other wavelet families \cite{dub92}. Therefore an orthogonal 8-tap Daubechies filter \cite{dub90}  in a tree structure decomposition is used to obtain the wavelet packets by expanding the basis having the most significant fractal signature, see Fig.~\ref{fig:wavelets}. This approach gives flexibility to finely tune the signal to the characteristic intrinsic properties of an image~\cite{wng08}.\\
			
\subsection{Multi-fractal textural model}
Fractals can be used in tissue characterization to describe irregular structures that exhibit semi self-similarity at different scales, and can further give an estimation of surface roughness (in our case of the RF envelope surface). There are several fractal models used to estimate the fractal dimension (FD); the FD can be estimated via the fractal Brownian motion (fBm) defined in (6) below, which is a non-stationary model known for its capability for describing random phenomena \cite{lop09}. Its statistical invariance to dilation, translation and rotation, can mitigate multiplicative speckle scale changes, making it a perfect candidate to be integrated with the Nakagami modeling and multi-resolution decomposition:

\begin{equation}
E\left(\Delta v \right) = K\Delta r^H
\end{equation}

\noindent where $E\left(\Delta v \right) = \left|q_i - p_j\right|, \!j = 1,\ldots, k$ is the mean absolute difference of voxel pairs $\Delta v$; $\Delta r = \sqrt{\sum^{n}_{i=1}\left(q_i - p_i\right)^2}$ where $n$ = 3 for 3-D space, is the voxel pair distances; $H$ is called the Hurst coefficient; and the constant $K$ $>$ 0.

\subsubsection{Fractal map estimation}

\begin{figure}
\begin{center}
\includegraphics[scale = 0.95]{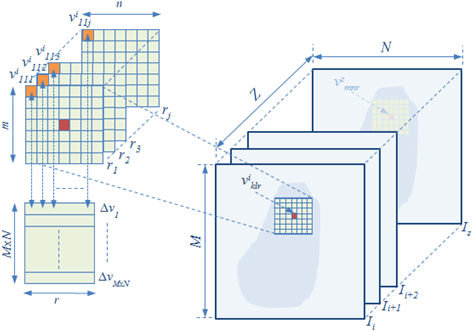}
\end{center}
\caption{Overcomplete multi-scale volumetric Nakagami tumor regions; the small $m \times n \times i$ voxel lattice centered on the localized voxel $v_{klr}$, where $k$, $l$, and $r$ are the voxel position on the lattice for scale $r$, is convolved with larger voxels up to $j$ resolution levels for estimation of the mean absolute difference of voxel pairs matrix.}
\label{fig:multiscale-regions}
\end{figure}

After application of the Daubechies 3D wavelet analysis, the roughness of each voxel lattice surface is determined via estimating its corresponding FD. The estimated voxel-by-voxel array of fractal dimensions for each voxel lattice, which we call a {\em fractal map}, provides a basis for characterizing the tissue and for building a {\em bag-of-words} of fractal features as a 3D feature descriptor. \\ 
   \indent A multi-dimensional matrix $N_{xyd}$ defined for each of the tumor voxels $v_{klr}$ is derived at different range scales $r$, such that the mean absolute difference of each voxel pair $\Delta v$ and for each voxel pair distances $\Delta r$ are estimated. Thereby the first dimension $d$ represents the voxel after it has been scaled once, and the second dimension represents the voxel at scale 2, and so on until the highest scale $j$ is reached.
	
\begin{equation}
N_{xyd} =	\left(\begin{array} {ccccc}
v^{i}_{11d} & v^{i}_{12d} & \cdots & \cdots & v^{i}_{1Nd}		\\
v^{i}_{21d} & v^{i}_{22d} & \cdots & \cdots & v^{i}_{2Nd}		\\
\vdots & \vdots & \hspace{-0.05\textwidth}\ddots &  & \vdots		\\
\vdots & \vdots &      &\ddots& \vdots		\\
v^{i}_{M1d} & v^{i}_{M2d} & \cdots & \cdots & v^{i}_{MNd}		\\
\end{array}\right)
\end{equation}

\noindent where $M$ and $N$ are the size of each ultrasound image slice and $d = 1, \ldots , j$ is the resolution limits of matrix $N_{xyd}$ which represents the mean absolute intensity difference to center voxels, and $i$ stands for Nakagami shape $\mu$ and scale $\omega$ parametric images. Then each element from each array in $N_{xyd}$ is normalized after taking the logarithm and saved in a mean absolute difference row vector $\Delta \hat{v}$. That is, the first element in all arrays of $N_{xyd}$ will compose vector $\Delta \hat{v}_1$, and all second elements will compose vector $\Delta \hat{v}_2$, and so on as shown in (8). This process is illustrated in Fig.~\ref{fig:multiscale-regions}. 

\begin{equation}
\left(\begin{array} {c}
\Delta \hat{v}_1		\\
\Delta \hat{v}_2		\\
\Delta \hat{v}_3		\\
\vdots		\\
\Delta \hat{v}_{M \times N}		\\
\end{array}\right) =	log\left(\begin{array} {ccccc}
\frac{\Delta \hat{v}^{i}_{111}}{\left\|\Delta \hat{v}^{i}_{111}\right\|} & \frac{\Delta \hat{v}^{i}_{112}}{\left\|\Delta \hat{v}^{i}_{112}\right\|} & \cdots & \cdots & \frac{\Delta \hat{v}^{i}_{11j}}{\left\|\Delta \hat{v}^{i}_{11j}\right\|}		\\
\frac{\Delta \hat{v}^{i}_{121}}{\left\|\Delta \hat{v}^{i}_{121}\right\|} & \frac{\Delta \hat{v}^{i}_{122}}{\left\|\Delta \hat{v}^{i}_{122}\right\|} & \cdots & \cdots & \frac{\Delta \hat{v}^{i}_{12j}}{\left\|\Delta \hat{v}^{i}_{12j}\right\|}		\\
\vdots & \vdots & \hspace{-0.05\textwidth}\ddots &  & \vdots		\\
\vdots & \vdots &      &\ddots& \vdots		\\
\frac{\Delta \hat{v}^{i}_{MN1}}{\left\|\Delta \hat{v}^{i}_{MN1}\right\|} & \frac{\Delta \hat{v}^{i}_{MN2}}{\left\|\Delta \hat{v}^{i}_{MN2}\right\|} & \cdots & \cdots & \frac{\Delta \hat{v}^{i}_{MNj}}{\left\|\Delta \hat{v}^{i}_{MNj}\right\|}		\\
\end{array}\right)
\end{equation}

The slope -- which corresponds to the Hurst coefficient $H$ -- of the least square linear regression line of the log-log plot of $\Delta \hat{v}$ versus $\Delta \hat{r}$ can be determined by means of sums of squares as in (9).

\begin{equation}
S_{rr} = \sum^{j-1}_{i=1} \Delta \hat{r}^2_i - \frac{\left(\sum^{j-1}_{i=1}\Delta\hat{r}_i\right)^2}{j-1} , S_{rv} = \sum^{j-1}_{i=1}\sum^{j-1}_{k=1}\Delta \hat{r}_i \hat{v}_k - \frac{\left(\sum^{j-1}_{i=1} \Delta\hat{r}_i\right)\left(\sum^{j-1}_{k=1}\Delta\hat{v}_k\right)}{j-1}
\end{equation}

Finally, the slope of the linear regression line defines the textural fractal characteristics, which we call the \textit{fractal map} $\Im$:

\begin{equation}
\Im = 3 - \frac{S_{rv}}{S_{rr}} = \left(\begin{array} {ccccc}
H_{11} & H_{12} & \cdots & \cdots & H_{1N}		\\
H_{21} & H_{22} & \cdots & \cdots & H_{2N}		\\
\vdots & \vdots & \hspace{-0.05\textwidth}\ddots &  & \vdots		\\
\vdots & \vdots &      &\ddots& \vdots		\\
H_{M1} & H_{M2} & \cdots & \cdots & H_{MN}		\\
\end{array}\right)
\end{equation}

\subsubsection{Volume of interest refinement}

It is important to estimate the Nakagami model parameters with good accuracy, but still have a simple model that is easy to interpret. Estimation from small cuboids of interest can provide poor estimation of the Nakagami parameters \cite{lru14}. Larger volumes have more data points to fit allowing for averaging of random error, yet this might not be good for tumors with relatively small size. In order to balance the trade off, volume reconstruction was designed to eliminate 2D tumor slices with low information content in order to provide a good characterization of tumor textural patterns. 

In practice, as a tumor grows it tends to adopt a non-uniform shape. This will cause sections in the acquired tumor volume to have a relatively small area compared to the whole tumor, where the texture patterns within these small regions cannot be reliably extracted. Therefore removal of these small regions will not only assist in reducing irrelevant features, computational time and memory, but will also direct the efforts of the developed feature descriptor to focus on characterizing the tumor patterns provided in large volume. The selection of volume slices was performed such that $A_i > \tilde{A}_m$, where $A_i$ is the tumor area in slice $i = 1, \ldots, m, \ldots, z$, and $\tilde{A}_m$ is the slice $m$ with the median area size.

\begin{figure}
\begin{center}
\includegraphics[scale = 1]{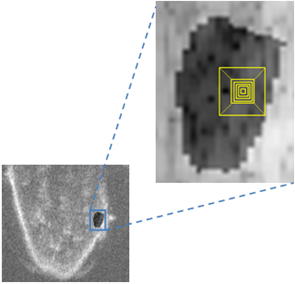}
\end{center}
\caption{Segmented liver tumor volume of interest; and the annotation squares in the enlarged image show the variation of voxel lattice size used in the experiments.}
\label{fig:voxel-lattice}
\end{figure}

    \indent Another important design decision is the selection of the size of the lattice utilized in Nakagami distribution estimation which is ideally performed automatically. To address this, a varying size voxel lattice was introduced as illustrated in Fig.~\ref{fig:voxel-lattice} to measure the goodness of fit for the estimated Nakagami model parameters. For error of fit we used the estimated root mean square error (RMSE)  between the MLE-estimated Nakagami values  $x_{mle}$ and the observed voxels values in each voxel lattice $x_v$ starting at size 2, as there would be no meaning if a lattice had a size of 1. The RMSE was thus defined as:
		
		\begin{equation}
		RMSE = \sqrt{\frac{\sum_{s=2}^{n}\left(x_{mle} - x_v\right)^2}{n} . }		
		\end{equation}
\\

\begin{figure}
\begin{center}
\includegraphics[scale = 0.85]{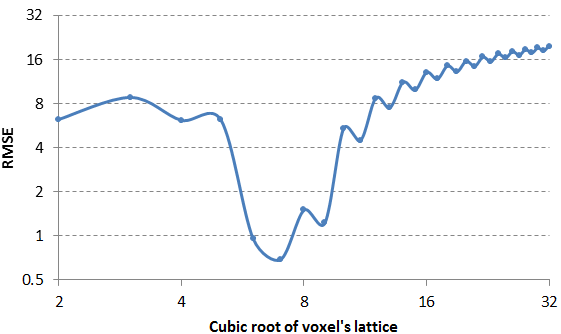}
\end{center}
\caption{Goodness of fit for optimizing the voxel lattice size utilized in the Nakagami distribution fitting.}
\label{fig:lattice-optimization}
\end{figure}

\indent Fig.~\ref{fig:lattice-optimization} is an empirical plot of the goodness of fit of the estimated Nakagami parameters versus voxel lattice size for a typical dataset. Sizes varying from 0.03 mm$^3$ to 6.60 mm$^3$ where used in the experiments. The RMSE oscillates as it reaches its minimum, recording a residual error of 0.68 at a lattice size of 7 voxels before the accuracy starts to decrease for larger sizes. Also in the process of generating the Nakagami parametric images and when the voxel lattice happen to be on the border, all voxels laying outside the volume of interest are eliminated from the calculations in order to discard any bias and to maintain a more credible estimation of the parameters. 
		
\subsubsection{Feature selection optimization}

The ultrasound texture fractal maps representing the FD voxel-based signature for the estimated Nakagami shape and scale parametric images are shown in Fig.~\ref{fig:voxel-characterization} (e) and (f), respectively. Various wavelet decomposition techniques apply the sub-bands' energy for decomposition which is susceptible to intensity variations in ultrasound images due to speckle; however, the local density function known as the FD, can instead overcome these local variations in voxel intensities as it gives a representation of texture surface roughness, and hence is employed for the multiresolution analysis. The fractal characteristics are estimated for all sub-bands at each level of the wavelet packet decomposition, where the FD is computed on a voxel-by-voxel basis to produce a fractal map $\Im$ for each sub-band i.e., each voxel in the fractal map has its own localized FD value estimated from its neighborhood as described in the previous section, where the rougher the surface the higher the FD values get, and vice versa. Features at boundaries are computed after assuming that each slice is mirror-like continually extended in both directions.  Specifically, the fractal features $f_{i,j}$ for a specific sub-band $j$ to a certain level of decomposition $i$ represent the average value of the generated $M \times N$ fractal image map $\Im$ of a volume of interest $k$ as defined in equation (12).
		
		\begin{equation}
		f_{i,j}^k
 = \frac{1}{MN}\sum_{r=1}^{MN}\Im_r		
		\end{equation}
		
This local estimation gives a more reliable estimation compared to a single global value. Finally, the optimized multi-fractal feature vector descriptor consists of all selected sub-band fractal feature signatures $f$ in each volume of interest $k$, expressed as $\Lambda = \bigcup_k \lambda_k$, where $\lambda_k = \left\{f^{1}_{1,1}, \ldots , f^{k}_{i,j}, \ldots , f^{z}_{m,n}\right\}$. 

In order to save processing time, the dimensionality of the extracted feature vector is reduced by applying a differential threshold which eliminates weak FD signatures. The threshold is defined by the condition $\bigl(\forall \bigl|f^{k}_{i+1,j} -$ $f^{k}_{i+1,j+1} \bigr|\in \Lambda \bigr) \leq D_f$ such that the FD signature absolute difference of the previous decomposition level $D_f = \left|f^{k}_{i,j} - f^{k}_{i,j+1}\right|$ is satisfied, then the decomposition should terminate. The new 3D Multi-fractal Nakagami Feature descriptor is abbreviated subsequently as MNF, and its estimation summarized in the following pseudo code:

\begin{algorithm} [!ht]
  \caption{Multifractal Nakagami Feature Descriptor Estimation}
  \begin{algorithmic}[1]
    \REQUIRE Volumetric ultrasound images \textit{\textbf{I}}, volume of interests from each volume \textit{\textbf{I}}: $I_i = \left\{\left(x_1,y_1,z_1,\ldots,x_{L_n},y_{L_n},z_{L_n}	 \right)\right\}$, segmented volume of interests: $\left\{V_s\right\}_{s=1}^{Z}$.
    \ENSURE multifractal feature descriptor $\left\{\Lambda_{f}^{(k)}\right\}$
    \FORALL {Segmented volumes of interest $V_1 \rightarrow V_Z$} 
		    \bigskip
				\STATE \COMMENT{\textbf{Step 1}} //\underline{subdivide each volume of interest $V_l$ into voxel lattices $v^i$.}
				\FORALL {Voxel lattices $v_{111}^{1} \rightarrow v_{mnj}^{z}$}
        \STATE Fit with a Nakagami distribution 
				$N(x|\mu,\omega)=2\left(\frac{\mu}{\omega}\right)^{\mu}\frac{1}{\Gamma\left(\mu\right)} x^{\left(2\mu-1\right)}$
				\bigskip
				\STATE \COMMENT {\textbf{Step 2}} //\underline{calculate Nakagami shape $\mu$ and scale $\omega$ parameters us-}
				\underline{ing maximum likelihood estimation as:}
				\STATEx $\hat{\theta}\left(v\right) = arg max_\theta \: D\left(\theta/v^1_{111},\ldots,v^z_{mnj}\right)$
				\STATEx where $\theta$ is a vector of parameters for the Nakagami distribution family $f\left(v^1_{111},\ldots,v^z_{mnj}/\theta\right)$
				\bigskip
				
				\STATE \COMMENT {\textbf{Step 3}}//\underline{construct Nakagami shape $N_\mu$ and scale $N_\omega$ parametric}
				\underline{array.}
				\ENDFOR
										
		\algstore{mnf_alg}
	\end{algorithmic}
\end{algorithm}
						
\begin{algorithm}
	\ContinuedFloat
	\caption{Multifractal Feature Descriptor Estimation (continued)}
		\begin{algorithmic} [1]
			\algrestore{mnf_alg}
					
			\ENDFOR
				\FORALL {voxels $v_{i}$ in $N_{\mu}$ and $N_{\omega}$}
				\bigskip
				\STATE \COMMENT {\textbf{Step 4}}//\underline{Perform Daubechies wavelet packet transform}
				\STATEx $W_\varphi \left(t_0, x, y, z\right) = \left({1}/{\sqrt{mnj}}\right) \sum^{j-1}_{r=0} \sum^{n-1}_{l=0} \sum^{m-1}_{k=0} v_{klr}\varphi_{klr}\left(t_0, x, y, z\right)$
		 \STATEx $W^i_\psi \left(t, x, y, z\right) = \left({1}/{\sqrt{mnj}}\right) \sum^{j-1}_{r=0} \sum^{n-1}_{l=0} \sum^{m-1}_{k=0} v_{klr}\psi^i_{klr}\left(t, x, y, z\right)$
				\bigskip

				\STATE \COMMENT {\textbf{Step 5}}//\underline{multifractal estimation and optimization phase}
		
		\FORALL {decomposition levels $i$}
				\FORALL {voxels $v_{mnr}^{i}$ in $W_{\varphi}$ and $W_{\psi}^{i}$}
						\FORALL {voxel pair distances $\Delta r$ in $W_{\varphi}$ and $W_{\psi}^{i}$}
						\STATE compute mean absolute difference $\Delta v$ of each voxel pair $q_i$, $p_i$
						\bigskip
						\STATE \COMMENT {\textbf{Step 6}}//\underline{construct a multidimensional volume of interest ma-} 
						\underline{-trix $N_{d}\left(x, y, d\right)$}
						\STATE normalize and take the logarithm $\Delta\hat{v} = log\left({\Delta v_{mnr}^{i}}/{\left\|\Delta v_{mnr}^{i}\right\|}\right)$, 
						\STATEx where $m$, $n$ and $r$ are the size of a voxel $i$ at a certain scale
						\STATE normalize voxel pairs distances $\Delta\hat{r}$ where $\Delta r = \sqrt{\sum^{n}_{i=1}\left(q_i - p_i\right)^2}$
						\STATE perform least square linear regression as 
						\STATEx $S_{rr} = \sum^{j-1}_{i=1} \Delta \hat{r}^2_i - {\left(\sum^{j-1}_{i=1}\Delta\hat{r}_i\right)^2}/{(j-1)}$ ,
						\STATEx $S_{rv} = \sum^{j-1}_{i=1}\sum^{j-1}_{k=1}\Delta \hat{r}_i \hat{v}_k - {\left(\sum^{j-1}_{i=1} \Delta\hat{r}_i\right)\left(\sum^{j-1}_{k=1}\Delta\hat{							v}_k\right)}/{(j-1)}$
						\STATE estimate the Hurst coefficient \textit{\textbf{H}} matrix which represents the slope $ H = \left(S_{rv}/S_{rr}\right)$
						\STATE estimate the fractal map  $\Im = 3 - H$
						\ENDFOR
				\ENDFOR
				\STATE extract mean fractal dimension $f_{i,j}^{k} \leftarrow \left(1/MN\right)\sum_{r=1}^{MN}\Im_r$
				\STATEx where $\lambda_k = \left\{f^{1}_{1,1}, \ldots , f^{k}_{i,j}, \ldots , f^{z}_{m,n}\right\}$
				\STATE construct feature descriptor from all wavelet sub-bands $\Lambda = \bigcup_k \lambda_k$
				\STATE $\left|f^{k}_{i+1,j} - f^{k}_{i+1,j+1} \right|$
				\bigskip
						\REPEAT
												
    	\algstore{mnf_alg2}
	\end{algorithmic}
\end{algorithm}
						
\begin{algorithm}
	\ContinuedFloat
	\caption{Multifractal Feature Descriptor Estimation (continued)}
		\begin{algorithmic} [1]
	    	\algrestore{mnf_alg2}
						
			\STATE \COMMENT {\textbf{Step 7}}//\underline{determine fractal absolute difference between decom-} 
						\underline{position level $i$ and subsequent level as:}
						\STATEx $D_{i}^{k} = \left|f^{k}_{i,j} - f^{k}_{i,j+1}\right|$,
						\STATEx $D_{i+1}^{k} = \left|f^{k}_{i+1,j} - f^{k}_{i+1,j+1}\right|$
						\UNTIL  $D_{i+1}^{k} \leq D_{i}^{k}$			
		\ENDFOR
		\ENDFOR
		\RETURN{} optimized multifractal feature vector: $\Lambda_{f}^{(k)} \leftarrow arg max\left(\lambda_{k}\right)$
  \end{algorithmic}
\end{algorithm}

\section{Experiments}

This section describes experiments on pre-clinical and clinical images to illustrate the new MNF algorithm and to compare its characterization performance with previous single scale methods. A tumor was classified as \textit{non-progressive} if categorized as partial response and \textit{progressive} if no change or disease demonstrated non-responsiveness. The response evaluation criteria in solid tumors (RECIST) was adopted to categorize the cases into progressive versus non-progressive \cite{eis09}. The baseline cross-sectional imaging was compared against those performed at the end of treatment according to the RECIST criteria to determine response to treatment for each target tumor.

\subsection{Data}

\subsubsection{Pre-clinical Data: Xenograft tumor imaging protocol}

RF ultrasound data was acquired using a diagnostic ultrasound system (z.one, Zonare Medical Systems, Mountain View, CA, USA) with a 10 MHz linear transducer and 50 MHz sampling. The output 2D image size was $20 \times 54$ mm with a resolution of $289 \times 648$ pixels. A total of 227 cross-sectional images of hind-leg xenograft tumors from 29 mice (20 progressive or stable disease and 9 non-progressive disease) were obtained with 1mm step-wise movement of the array mounted on a manual positioning device until the whole tumor volume was imaged (Fig.~\ref{fig:voxel-decomposition}). All studies were ethically approved and performed in line with UK Home Office regulations, and in accordance with personal and project licenses.

The 2D images were composed together to create a 3D ultrasound volume. In order to ensure that nearby healthy tissue is not included in tumor tissue characterization, two expert radiologists manually segmented each image prior to applying tissue characterization. The Nakagami distribution was fitted to the distributions in each voxel lattice and parametric volumes were generated for each tumor. 

The complete 3D RF ultrasound dataset along with a description of case categorization can be downloaded from the following weblink url: \url{https://ibme-web.eng.ox.ac.uk/livertumour}. An example of one of the cases presented as an animated GIF for the fractal slice maps and a video of the corresponding fractal volume map can be found with the dataset. 


\subsubsection{Clinical Data: Clinical study imaging protocol}

Cross-sectional images of liver tumors undergoing chemotherapy treatment obtained as part of an ethically approved prospective study was used to validate our proposed technique.  A total of 394 cross-sectional images (186 from tumors demonstrating partial response categorized as non-progressive, and 208 from tumors demonstrating progressive disease categorized as progressive) were obtained using a diagnostic ultrasound system (z.one, Zonare Medical Systems, Mountain View, CA, USA) with a 4 MHz curvilinear transducer and 11 MHz sampling. Each dataset was acquired prior to commencement of chemotherapy. Response to treatment was determined based on conventional Computed Tomography follow up imaging as part of the patient standard clinical care according to the RECIST criteria \cite{eis09}. \\
\indent The transducer beam was initially directed through the target liver tumor in the intercostal imaging plane. Patients were asked to maintain breath hold inspiration, in order to stabilize the tumor target during image acquisition. Using a smooth movement of approximately constant speed, the ultrasound probe was angled whilst maintaining a skin contact position in a cranial to caudal direction to capture sequential 2D cross-sectional images of the target liver tumor. Each output 2D image size was $65 \times 160$ mm with a resolution of $225 \times 968$ pixels. Similar to the xenograft tumor dataset, the 2D images were composed together to create a 3D ultrasound volume for each target tumor. The acquisition was repeated in a similar fashion three times at each time point. Manual segmentation of the liver tumor was also performed in a similar fashion prior to image texture analysis.

\begin{figure} [ht]
\includegraphics[width=\columnwidth]{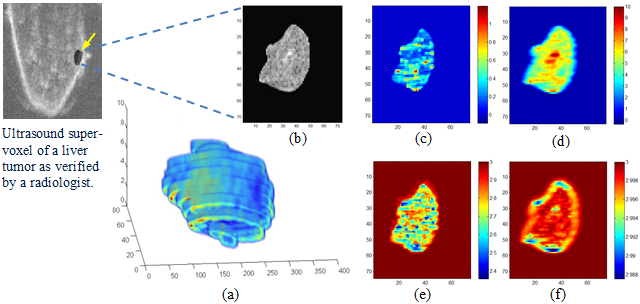}
\caption{Example of a voxel-based tissue characterization for a non-progressive liver tumor case. The tumor 3D volume is reconstructed in (a), and the B-mode middle slice (b) after transforming using the MNF algorithm is shown in (c-f). The Nakagami shape and scale parametric voxels (c) and (d) and the corresponding multi-resolution fractal slice maps (e) and (f) illustrates how the case responds to chemotherapy treatment -- the blue color regions in (e) and (f) which correspond with the RECIST criteria.}
\label{fig:voxel-characterization}
\end{figure}

\begin{figure} [ht]
\begin{center}
\includegraphics[scale = 1]{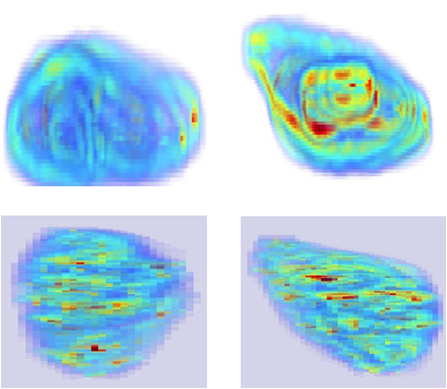}
\end{center}
\caption{Fractal volume maps: Volumetric rendering of Nakagami parametric scale (first row) and shape (second row) for a progressive and non-progressive liver tumor volume, respectively.}
\label{fig:volume-maps}
\end{figure}

		\begin{figure} [ht]
\includegraphics[width=\columnwidth]{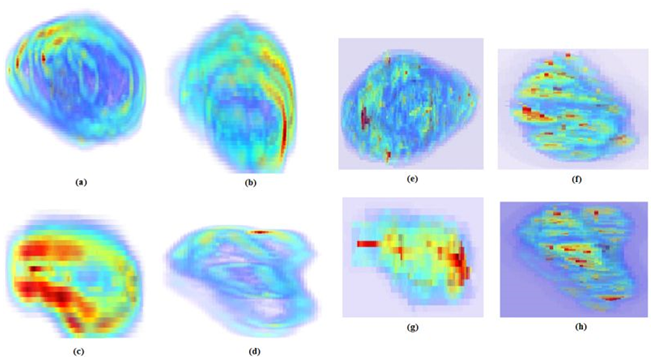}
\caption{Pairwise horizontal comparison of texture-based volume rendering of Nakagami scale (a-d) and shape (e-h) multi-fractal volume maps. (a) \& (b) are an example of a non-progressive case in pre and post-chemotherapy treatment, and the (c) and (d) are for a progressive case in pre and post-chemotherapy treatment, respectively; (e) and (f), and (g) and (h) are the corresponding volumetric Nakagami shape cases. Red color labels indicate low local fractal dimension or low-activity regions which correspond to necrotic regions according to RECIST criteria. In first row, it is noticed that the spread of the red voxels has increased in post-treatment as compared to pre-treatment, and vice versa in the second row.}
\label{fig:therapeutic-comparison}
\end{figure}

\begin{figure} [ht]
\includegraphics[scale = 0.85]{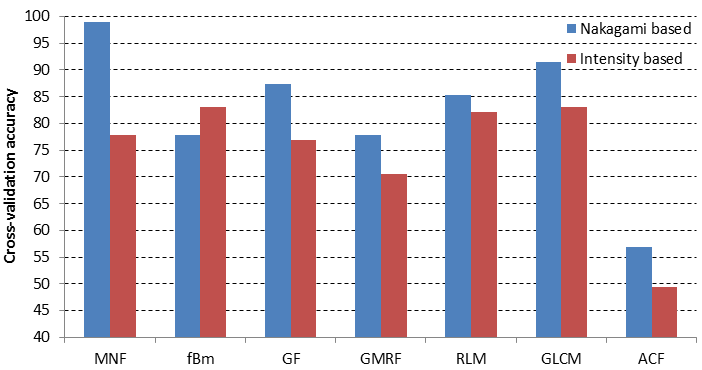}
\caption{Performance comparison for the MNF method against the filter-based Gabor filter (GF), model-based fractional Brownian motion (fBm) and Gaussian Markov random field (GMRF), and statistical-based gray-level co-occurrence matrix (GLCM), run-length matrix (RLM), and autocovariance function (ACF) texture analysis methods. The blue columns represent the operation of the texture descriptors on volumetric Nakagami parametric volume of interests, while the red columns are results from conventional ultrasonic intensity B-mode volume of interests.}
\label{fig:performance-comparison}
\end{figure}

\subsection{Fractal maps}

Fig.~\ref{fig:voxel-characterization} shows the new parametric mapping on an example case. The various scattering conditions related to tissue characteristics within the tumor texture are shown using color mapping, where the various intensity distributions corresponding to the local concentration of scatterers vary from pre-Rician (0 $<$ $\mu$ $<$ 0.5), generalized Rician ($\mu$ = 0.5), pre-Rayleigh (0.5 $<$ $\mu$ $<$ 1), Rayleigh ($\mu$ = 1), post-Rayleigh ($\mu$ $>$ 1) as illustrated in Fig.~\ref{fig:voxel-characterization} (c), and from low (0 $<$ $\omega$ $<$ 3), mid (3 $\leq$ $\omega$ $<$ 7), and high ($\omega$ $>$ 7) for local backscattered energy as in Fig.~\ref{fig:voxel-characterization} (d). The MNF \textit{fractal slice maps} -- which correspond to each slice in the reconstructed volume -- shown in Fig.~\ref{fig:voxel-characterization} (e) and (f) correspond to the Nakagami shape and scale voxels of the mid-slice, respectively. Regions with different texture properties (smoother or with lower local fractal dimension values) become more apparent compared to the Nakagami parametric voxels. Another interesting point is that the scale fractal map highlights the low activity regions near to the edge of the tumor, while the shape fractal map complements the characterization by highlighting low activity regions belonging to the inner part of the tumor tissue texture. A holistic view of the overall progression or regression of tumor spread can be effectively revealed via the \textit{fractal volume maps} where low activity regions which correspond to necrotic tissue are labeled by a dark red color as illustrated in Fig.~\ref{fig:volume-maps}. Also a comparison between a non-progressive and progressive case and for pre and post-chemotherapy treatment is shown in Fig.~\ref{fig:therapeutic-comparison}.

\subsection{Statistical analysis}

Since our primary concern is to  demonstrate the texture expressiveness of the new multi-fractal feature descriptor in this subsection we describe experiments conducted to compare the discriminative power of the MNF descriptor with established features for mass classification. To do this we have chosen to use a na\"{\i}ve” Bayesian classifier,  although SVM or random forests might also have been used. 

Specifically, we consider the MNF  $\Lambda^{(k)}$ estimated over multiple scales in each case $k$ in which there are $\left\vert{\Lambda^{(k)}}\right\vert = 8 \times i$ features per voxel, where $i$ is the number of decomposition levels estimated adaptively (see step 7 in Algorithm 1). This feature vector was fed into a simple na\"{\i}ve” Bayesian classifier (nBC) determine performance of classifying progressive versus non-progressive cases. 

Fig.~\ref{fig:performance-comparison} summarizes classification results for the pre-clinical dataset using the proposed MNF features with results from six other classic filter, model and statistical-based texture analysis methods and for B-mode intensity and Nakagami-based volumes of interest. The compared texture analysis methods are: Gabor filter (GF), fractional Brownian motion (fBm), Gaussian Markov random field (GMRF), gray-level co-occurrence matrix (GLCM), run-length matrix (RLM), and autocovariance function (ACF). Details of the extracted features can be found in Table~\ref{table:Extracted_features}, and a leave-one-out validation approach was employed. MNF-based performance gave the best overall cross-validation accuracy of 98.95\%. Also the Wilcoxon Signed-Rank test on paired accuracy in Nakagami and intensity-based of each subject for both two-class classification shows that there is a significant difference ($p < 0.05$).

\begin{table} [!ht]
\centering
\captionsetup{justification=centering}
\caption{Features extracted from the comparative texture analysis methods in Fig.~\ref{fig:performance-comparison}}
\label{table:Extracted_features}
\begin{tabular}{p{2cm}p{9cm}}
\hline \hline
Method & \multicolumn{1}{c}{Texture features}  \\ \hline
GF & Energy of each magnitude response for five radial frequencies $\left(\sqrt{2}/2^{6}, \sqrt{2}/2^{5}, \sqrt{2}/2^{4}, \sqrt{2}/2^{3}, \sqrt{2}/2^{2}\right)$ with 4 orientations 0$^{\circ}$,45$^{\circ}$,90$^{\circ}$ \& 135$^{\circ}$  \\
GMRF & Seven features estimated from a third order Markov neighborhood model  \\
fBm & Mean, variance, lacunarity, skewness and kurtosis derived from the generated FD image \\
GLCM & Contrast, correlation, energy, entropy, homogeneity, dissimilarity, inverse difference momentum, maximum probability statistical features derived in the 0$^{\circ}$,45$^{\circ}$,90$^{\circ}$ \& 135$^{\circ}$ directions \\
RLM & Short run emphasis, long run emphasis, gray level non-uniformity, run length non-uniformity and run percentage statistical features derived in the 0$^{\circ}$,45$^{\circ}$,90$^{\circ}$ \& 135$^{\circ}$ directions \\
ACF & Peaks of the horizontal and vertical margins values and associated exponential fittings of the ACF  \\ \hline \hline
\end{tabular}
\end{table}

\subsection{Clinical application}

In order to demonstrate the applicability of the MNF algorithm for analysis of data acquired to clinical protocol the new method was applied to a clinical liver tumor dataset. In this case, RF ultrasound data is acquired in a fan-like scanning protocol (cf. the pre-clinical dataset was acquired using a linear transducer) in which a series of 2D images are collected as the transducer is tilted and then reconstructed into a 3D image. \\
\indent Table~\ref{table:MNF_classification} summarizes the classification performance for the clinical dataset following the same classifier design as in section 4.3. The results show a good classification accuracy of 92.90\% using a leave-one-out  cross-validation approach, and a 92.01\% $\pm$ 0.50 and 92.60 $\pm$ 0.30 for 5-fold and 10-fold cross-validation (results are the mean $\pm$ standard deviation of the performance over 60 runs). The texture descriptor was also compared against the RF backscatter signal using the localized voxel-based Nakagami parameters without generating the fractal volume maps (see Table~\ref{table:Nakagami_classification}), and by only deriving the fractal maps directly from the intensity (B-mode) images as well (see Table~\ref{table:B-mode_classification}).
	
\begin{table} [!ht]
\centering
\captionsetup{justification=centering}
\caption{Detailed Classification Performance for the 3D Clinical RF Ultrasound Liver Tumor Test Set Using the MNF Algorithm}
\label{table:MNF_classification}
\begin{tabular}{lccc}
\hline \hline
\multirow{2}{*}{\begin{tabular}[c]{@{}l@{}}Classification \\ Performance\end{tabular}} & \multicolumn{3}{c}{Cross-validation}   \\ \cline{2-4}
 & loo & 5-fold & 10-fold   \\ \hline
Recall & 0.935 & $0.92 \pm 0.919$ & $0.921 \pm 0.931$   \\
FP rate & 0.065 & $0.08 \pm 0.080$ & $0.069 \pm 0.079$  \\
Accuracy & 0.929 & $0.92 \pm 0.005$ & $0.926 \pm 0.003$  \\
Precision & 0.941 & $0.93 \pm 0.911$ & $0.937 \pm 0.914$  \\
F-measure & 0.928 & $0.92 \pm 0.005$ & $0.925 \pm 0.003$  \\
J-Index & 0.929 & $0.92 \pm 0.005$ & $0.926 \pm 0.003$  \\
Dice SC & 0.963 & $0.96 \pm 0.003$ & $0.961 \pm 0.002$  \\
ROC Area & 0.929 & $0.92 \pm 0.006$ & $0.926 \pm 0.003$ \\ \hline \hline
\end{tabular}
\end{table}

\begin{table} [!ht]
\centering
\captionsetup{justification=centering}
\caption{Detailed Classification Performance for the 3D Clinical RF Ultrasound Liver Tumor Test Set using only the Nakagami Parameters}
\label{table:Nakagami_classification}
\begin{tabular}{lccc}
\hline \hline
\multirow{2}{*}{\begin{tabular}[c]{@{}l@{}}Classification \\ Performance\end{tabular}} & \multicolumn{3}{c}{Cross-validation}   \\ \cline{2-4}
 & loo & 5-fold & 10-fold   \\ \hline
Recall & 0.715 & $0.73 \pm 0.492$ & $0.72 \pm 0.491$   \\
FP rate & 0.514 & $0.51 \pm 0.272$ & $0.51 \pm 0.276$  \\
Accuracy & 0.594 & $0.60 \pm 0.007$ & $0.60 \pm 0.006$  \\
Precision & 0.656 & $0.56 \pm 0.670$ & $0.56 \pm 0.666$  \\
F-measure & 0.595 & $0.60 \pm 0.007$ & $0.60 \pm 0.006$  \\
J-Index & 0.594 & $0.60 \pm 0.007$ & $0.60 \pm 0.006$  \\
Dice SC & 0.745 & $0.75 \pm 0.005$ & $0.75 \pm 0.005$  \\
ROC Area & 0.600 & $0.61 \pm 0.007$ & $0.61 \pm 0.006$ \\ \hline \hline
\end{tabular}
\end{table}

\begin{table} [!ht]
\centering
\captionsetup{justification=centering}
\caption{Detailed Classification Performance Using B-Mode Images of the 3D Clinical Ultrasound Liver Tumor Test Set}
\label{table:B-mode_classification}
\begin{tabular}{lccc}
\hline \hline
\multirow{2}{*}{\begin{tabular}[c]{@{}l@{}}Classification \\ Performance\end{tabular}} & \multicolumn{3}{c}{Cross-validation}   \\ \cline{2-4}
 & loo & 5-fold & 10-fold   \\ \hline
Recall & 0.823 & $0.82 \pm 0.870$ & $0.82 \pm 0.868$   \\
FP rate & 0.135 & $0.13 \pm 0.182$ & $0.13 \pm 0.180$  \\
Accuracy & 0.845 & $0.85 \pm 0.005$ & $0.85 \pm 0.004$  \\
Precision & 0.845 & $0.85 \pm 0.842$ & $0.85 \pm 0.844$  \\
F-measure & 0.845 & $0.85 \pm 0.005$ & $0.85 \pm 0.004$  \\
J-Index & 0.845 & $0.85 \pm 0.005$ & $0.85 \pm 0.004$  \\
Dice SC & 0.916 & $0.92 \pm 0.003$ & $0.92 \pm 0.002$  \\
ROC Area & 0.844 & $0.84 \pm 0.005$ & $0.84 \pm 0.004$ \\ \hline \hline
\end{tabular}
\end{table}

\section{Discussion}

An outstanding challenge in medical ultrasound image analysis addressed in the paper is to provide an efficient characterization of subtle speckle textural changes or intra-heterogeneity within tumor tissue. The low signal to noise ratio and imaging artifacts frequently present in clinical ultrasound images make extracting such diagnostically useful information hard. In this section we focus on discussing three main contributions of our work and their importance.

{\em 1) Features of the method which make it novel and how this makes a difference:}  We have proposed a novel and meaningful fractal feature vector representation of ultrasonic signal characterization across spatial scales. This uses a multi-scale analysis where the voxel lattice is optimized to tumor size.  We have also shown that performing the estimation in a volumetric fashion improves classification accuracy.

\indent The proposed MNF approach  simplifies analysis of the higher order statistics of the speckle texture via investigating different sub-bands at various decomposition levels using wavelets that tend to exhibit fractal characteristics. The tailored spatial-frequency localization provided via the Daubechies wavelet can facilitate the subsequent measurement of (signal) surface roughness while simultaneously filtering out irrelevant speckle features. The resulting fractal features quantify the speckle texture. Unlike conventional sub-band energy decomposition, sub-resolution level probing via the fBm (which satisfies an affine intensity invariance property) is not susceptible to sudden changes in the speckle intensity spatial frequency distribution.

{\em 2) Significance of the MNF descriptor to describe cancer morphology at the image level:} The results in section 4 are very interesting as they suggest that the MNF representation provides a new way to visualise cancer morphology using an ultrasound-based descriptor, and specifically to partition cancerous masses into necrotic and non-necrotic areas which has potential clinical utility. 

In particular, the sparse blue colored clusters -- which correspond to the low fractal dimension values shown in Fig.~\ref{fig:voxel-characterization} (e) and (f) -- appear to correspond to necrotic regions in the tumor that are beginning to respond to chemotherapy treatment as defined by the RECIST criteria. This can be explained by first noting that tumorous tissue tends to have a "rougher" appearance than non-tumorous tissue due to the chaotic way that tumors  build their network of new blood vessels. These angiogenesis networks tend to be leaky and disorganized, unlike blood vessel vasculature in normal tissue. This heterogeneity introduces a degree of randomness in appearance – or ``roughness''- and gives a higher fractal dimension value compared to necrotic tissue. 

    Necrotic regions have different echogenicity characteristics \cite{sdg13}. In a necrotic region there is no cell growth and hence could be quantified if investigated at the appropriate analysis scale. The fractal volume maps reveal some of the intra-heterogeneity regions that tend to have a different textural characteristics to that of tumor tissue. These observations correspond to the dark red voxels in Fig.~\ref{fig:volume-maps} and in Fig.~\ref{fig:therapeutic-comparison} in the post-chemotherapy case. The red voxels in the pre-treated tumors in Fig.~\ref{fig:therapeutic-comparison} (a) and (c) represent regions with low activity, i.e. non-aggressive, with potential to become aggressive after treatment. We consider these red voxels as suspicious regions within the tumour and not as active as the rest of the malignant tumour tissue. From a pattern recognition perspective, the red voxels are closer in terms of their surface roughness characteristics to necrotic (i.e. low fractal dimension values) rather than aggressive regions (i.e. high fractal dimension values). However, at the pre-treatment stage, the nature of these regions is not yet confirmed since the tumor has not been subjected to chemotherapy treatment. The different shades of red in the pre-treatment figures reflect the varying degree of low activity that exists in the tumor. Subsequently, and after the first session of chemotherapy treatment, we notice that increase in these low activity regions in the non-progressive tumor of Fig.~\ref{fig:therapeutic-comparison} (b). Since the tumor has responded to treatment, we can now be confident that the aforementioned low activity regions in fact refers to necrotic regions. Conversely, the red voxels seen on the pre-treatment images for the tumor that progressed post chemotherapy nearly disappears on the post-treatment images of Fig.~\ref{fig:therapeutic-comparison}, thus suggesting that they have now become aggressive resulting in progression of the tumor. 

\indent We also observe that the scale and shape fractal maps tend to complement each other, in the sense of highlighting different aspects of the analyzed speckle texture pattern. This is evident when examining the Nakagami shape and scale parametric voxels of the non-progressive case shown in Fig.~\ref{fig:voxel-characterization}. Here we observe that a number of necrotic regions -- highlighted in red in Fig.~\ref{fig:voxel-characterization} (c) and (d) - become more apparent in the corresponding fractal maps of Fig.~\ref{fig:voxel-characterization} (e) and (f). The scale fractal map highlights low intra-heterogeneity regions on the outer surface and near to the edge of the tumor. The shape fractal map complements this by revealing most of these low intra-heterogeneity regions belonging to the inner part of the tumor.

{\em 3) Expressiveness of MNF as a mass characterization feature:} Results in Fig.~\ref{fig:performance-comparison} for the pre-clinical cases show that the MNF performed best based on Nakagami-based parametric images reporting an accuracy of 98.95\%, while the fBm worked well on traditional B-mode intensity images with an accuracy of 83.16\%. Similarly for the clinical cases, where under all classification performance metrics by comparing Table~\ref{table:MNF_classification}, ~\ref{table:Nakagami_classification}, and ~\ref{table:B-mode_classification}, we can see that the MNF algorithm outperformed both using the Nakagami parameters alone and when texture analysis was applied to the B-mode images as well.
Moreover, we found that using a 2D single slice gave a lower accuracy of 74.74\%  compared to a volumetric analysis (98.95\%). Furthermore, combining the texture-based multiresolution fractal features extracted from the Nakagami maps via the MNF algorithm with the features extracted from the conventional B-mode intensity images resulted in 4.1\% degradation in the overall accuracy, as compared if the MNF method was employed alone. Note that in part due the limited data available in the pilot work, we have not looked at the benefits from a mass classification perspective of combining the MNF descriptor with other ultrasound tissue characterization parameters or image texture features which would be a natural topic to explore in the future. 

\indent Finally, we comment on three limitations of the current research which can be translated into opportunities for future investigation. 
Firstly, we are currently relying on the consensus of two radiologists to provide the gold standard and training data. Although we have shown good results with our current strategy, of possible concern is that the training samples may be mislabelled. This would reduce the accuracy of the results. Future work might look at the significance of this and strategies for mitigation.  
Secondly, fatty livers may result in attenuation of tissue properties and it would be interesting to investigate how this affects MNF classification accuracy. 
Thirdly,  RF characteristics (and hence speckle appearance) tends to differ between ultrasound devices. It would be interesting to investigate whether a training set from one ultrasound scanner can be used for classification of images from a different scanner or results are scanner specific.

\section{Conclusion}

A new approach for assessing tumor heterogeneity via 3D multi-fractal multi-scale Nakagami-based feature modeling has been presented which we believe is the first work to consider intra-heterogeneity quantification of a cancerous mass. We estimated volumetric Nakagami shape and scale parameters from which the novel fractal descriptor is estimated. Future work will investigate the use of the method for both staging liver tumors and in longitudinal analysis as an image-based biomarker of tumor growth and therapeutic response.

\section*{Acknowledgment}

This work was support by the Engineering and Physical Sciences Research Council and Wellcome Trust Grant WT 088877/Z/09/Z. The authors would like to thank the anonymous reviewers for their constructive comments and suggestions to improve the quality of the paper.

\section*{References}

	\bibliographystyle{elsarticle-num}
  \bibliography{MIA}

\end{document}